\documentclass{article}

\usepackage[preprint]{neurips_2025}

\usepackage[utf8]{inputenc} 
\usepackage[T1]{fontenc}    
\usepackage{hyperref}       
\usepackage{url}            
\usepackage{booktabs}       
\usepackage{amsfonts}       
\usepackage{nicefrac}       
\usepackage{microtype}      
\usepackage[pdftex]{graphicx}
\usepackage{pifont}
\usepackage{caption}
\usepackage{subcaption}
\usepackage{amsmath}
\usepackage[dvipsnames]{xcolor}
\usepackage{multirow}

\newcommand{\ours}{\textsc{HAMburger}}

\title{\ours{}: Accelerating LLM Inference via Token Smashing}

\author{%
    Jingyu Liu \\
    Department of Computer Science\\
    University of Chicago \\
    Chicago, IL 60637 \\
    \texttt{jingyu6@uchicago.edu} \\
    \And
    Ce Zhang \\
    Department of Computer Science\\
    University of Chicago \\
    Chicago, IL 60637 \\
    \texttt{cez@uchicago.edu}
}

\begin{document}

\maketitle


\begin{abstract}
The growing demand for efficient Large Language Model (LLM) inference requires a holistic optimization on algorithms, systems, and hardware. However, very few works have fundamentally changed the generation pattern: each token needs one forward pass and one KV cache. This can be sub-optimal because we found that LLMs are extremely capable of self-identifying the exact dose of information that a single KV cache can store, and many tokens can be generated confidently without global context. Based on this insight, we introduce \ours{}~\footnote{Project page: \url{https://github.com/Jingyu6/hamburger}}, a \textbf{H}ierarchically \textbf{A}uto-regressive \textbf{M}odel that redefines resource allocation in LLMs by moving beyond uniform computation and storage per token during inference. Stacking a compositional embedder and a micro-step decoder in between a base LLM, \ours{} smashes multiple tokens into a single KV and generates several tokens per step. Additionally, \ours{} functions as a speculative decoding framework where it can blindly trust self-drafted tokens. As a result, \ours{} shifts the growth of KV cache and forward FLOPs from linear to sub-linear with respect to output length, and adjusts its inference speed based on query perplexity and output structure. Extensive evaluations show that \ours{} reduces the KV cache computation by up to 2$\times$ and achieves up to 2$\times$ TPS, while maintaining quality in both short- and long-context tasks. Our method explores an extremely challenging inference regime that requires both computation- and memory-efficiency with a hardware-agnostic design. 
\end{abstract}

\section{Introduction}

\begin{figure}[ht]
\centering
\vspace{-1cm}
\makebox[\textwidth][c]{\includegraphics[width=\textwidth]{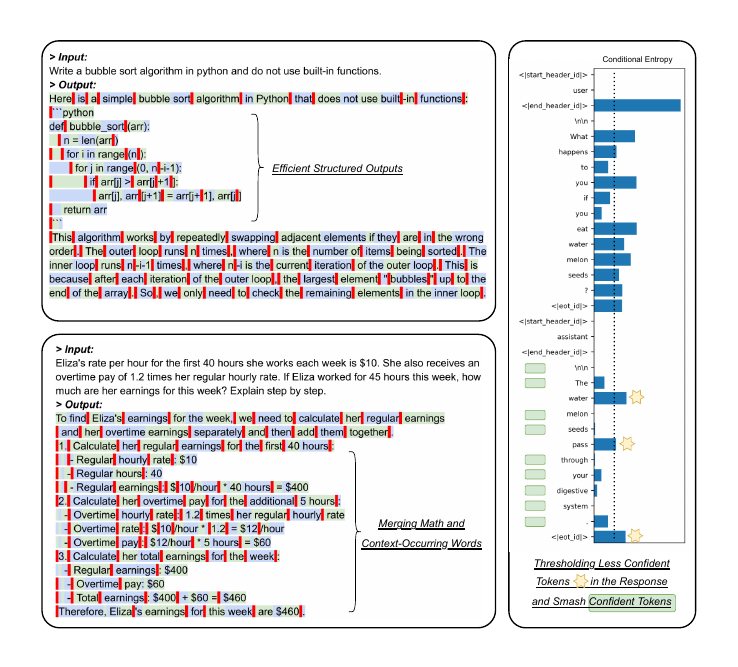}}
\vspace{-1cm}
\caption{How \textbf{\ours{} Merges Tokens}: We showcase two examples of dynamically generating a ``unit dose of information'' per step instead of a fixed single token (left). The output tokens from the static vocabulary are distinguished by alternating blues and greens. The red dividers separate groups of tokens that are \textit{predicted with a single macro-step} by \ours{}. During data pre-processing, we rely on model's own knowledge (i.e., conditional entropy) for segmentation (right). }
\label{fig:gen}
\vspace{-0.5cm}
\end{figure}

Large Language Models (LLMs), such as GPT-4~\cite{openai2024gpt4technicalreport}, Llama~\cite{grattafiori2024llama3herdmodels}, Claude~\cite{claude2025}, Qwen~\cite{qwen2025qwen25technicalreport}, and DeepSeek~\cite{deepseekai2025deepseekv3technicalreport}, have transformed natural language processing, enabling remarkable advancements in tasks ranging from text generation to complex reasoning~\cite{liang2023holisticevaluationlanguagemodels, feng2024faragillmsneed}. However, their computational demands, particularly during decoding, are still limiting their ubiquitous applications that require scalability and efficiency~\cite{miao2023efficientgenerativelargelanguage}. A critical issue is the linear growth of the Key-Value (KV) cache size and computation with each generated token, a direct result from uniform allocation of floating-point operations (FLOPs) across tokens, regardless of their computational necessity. Additionally, while speculative decoding~\cite{leviathan2023fastinferencetransformersspeculative} has emerged as a successful solution to reduce decoding latency in many cases, several limitations still remain, including the need for a separate drafting model with alignment in knowledge and tokenization, reduced performance with smaller base models, inefficiencies from verification with large batch sizes, and deployment complexities. 

Addressing these challenges is fundamental for the advancement of LLM inference efficiency for many generation-intensive applications. The central question we aim to answer in this work is whether we can develop a model that can dynamically adjust the resource allocation based on token properties. This will break the assumed unit of information defined by the tokenizer and create a virtual vocabulary that is determined by the capacity of a single KV cache. 

Our key insight is that \textit{LLMs are extremely capable of assessing a token's information, and not all tokens need global context for accurate generation}~\cite{li2024chunkdistilledlanguagemodeling, liu2025speculativeprefillturbochargingttft}. To take advantage of our findings, we propose \ours{}, a \textbf{H}ierachically \textbf{A}uto-regressive \textbf{M}odel that can generate multiple tokens dynamically based on its own knowledge per forward, which will then be smashed into a single KV cache for the next step. \ours{} stacks a base LLM with a relative-position-aware compositional embedder before and a local micro-step decoder after it, which handles token fusion and multi-token generation respectively. Our approach can also be framed as a self-speculative decoding framework where the drafting overhead is constant w.r.t. the context and all drafted tokens are blindly trusted with a single forward verification. 

We comprehensively evaluated our instruction fine-tuned model on both short- and long-context tasks and show that \ours{} can achieve up to 2$\times$ reduction in KV cache \textit{computation} (hence storage) and up to 2 $\times$ TPS improvement while maintaining or even surpassing the base model. Furthermore, \ours{} reduces the linear growth rate to sub-linear, paving the road for future generation-intensive applications. 
\section{Related Work}

\begin{table}[ht]
\centering
\caption{\textbf{Model Comparison}: We highlight the uniqueness of \ours{} by comparing several models, all of which attempt to accelerate decoding by ``predicting'' more than one token per step. We contrast them in terms of the vocabulary space as well as whether they can take dynamic inputs / outputs, need extra computation for verification, and can compress KV cache for efficiency. }
\label{tab:compare}
\vspace{0.2cm}
\begin{tabular}{lcccc}
\toprule
\textbf{Model Name} & \textbf{Vocab} & \textbf{Dynamic Gen} & \textbf{Verification} & \textbf{Compressed KV} \\
\midrule
\midrule
MegaByte~\cite{yu2023megabytepredictingmillionbytesequences} & Bytes & \ding{55} & \ding{55} & \ding{51} \\
Byte Latent Transformer~\cite{pagnoni2024bytelatenttransformerpatches} & Bytes & \ding{51} & \ding{55} & \ding{51} \\
Large Concept Models~\cite{lcmteam2024largeconceptmodelslanguage} & Concept & \ding{51} & \ding{55} & N/A \\
Spec Decoding Variants~\cite{leviathan2023fastinferencetransformersspeculative} & Tokens & N/A & \ding{51} & \ding{55} \\
\midrule
\ours{} (ours) & Tokens & \ding{51} & \ding{55} & \ding{51} \\
\bottomrule
\end{tabular}
\end{table}

Optimizing inference decoding has been a central question for LLMs, the techniques of which cover kernal optimization, speculative decoding, KV cache compression, and etc~\cite{miao2023efficientgenerativelargelanguage}. We review several major categories that share some similarity to \ours{} and we contrast each one of them to illustrate the uniqueness of our approach (comparison overview in Table~\ref{tab:compare}). 

\subsection{Multi-step Decoding}

Byte Latent Transformer (BLT)~\cite{pagnoni2024bytelatenttransformerpatches} takes and outputs a patch of bytes per step, with the help of a separate entropy model for segmenting the data. Despite showing promising scaling properties, BLT has several practical limitations: 1) a separate model needs to be trained and deployed and 2) using bytes as the granularity is not as efficient with the same model size. \ours{} addresses the problem by self-deciding how to segment the patches and generate with tokens. MegaByte~\cite{yu2023megabytepredictingmillionbytesequences} and Block Transformer~\cite{ho2024block} remove the entropy model but at a cost of fixing the patch size, which can be less flexible compared to \ours{} as the information per unit can vary drastically. Moving to token level, Large Concept Models (LCM)~\cite{chen2021evaluatinglargelanguagemodels} differs from standard LLMs in the representation space and diversity of modality. However, the method is still in its experimental phase and has not yet demonstrated their quality and real inference speedup in a wider range of tasks as ours. To our best knowledge, \ours{} is the first standalone multi-token prediction model that achieves competitive performance and better inference efficiency than the SOTA transformer LLMs. 

\subsection{Speculative Decoding}

Speculative Decoding~\cite{chen2023acceleratinglargelanguagemodel} trades off parallelled computation for memory during the decoding phase, which has been proven extremely effective with lossless quality~\cite{xia2024unlockingefficiencylargelanguage}. Many prior works attempt to improve the drafting efficiency by replacing linear prediction with tree drafting~\cite{xiong2024dyspecfasterspeculativedecoding, yao2025deftdecodingflashtreeattention, zhao2024lookaheadinferenceaccelerationframework}. A canonical example of these is Medusa~\cite{cai2024medusasimplellminference} which uses grafted heads to construct a draft token tree and verify with sparse attention mask. On top of that, Multi Token Prediction (MTP) from DeepSeek-V3~\cite{deepseekai2025deepseekv3technicalreport} takes advantage of the model training process for better alignment. EAGLE series~\cite{li2025eaglespeculativesamplingrequires, li2024eagle2fasterinferencelanguage} propose to use the internal features from the base model to improve the acceptance rate, which is the core factor directly deciding the performance of speculative sampling methods. Chunk-Distilled Language Modeling~\cite{li2024chunkdistilledlanguagemodeling} applies a carefully designed retrieval module as a drafter, providing fast adaptation to new domain and disentangling the draft model (and tokenizer). Despite impressive success, there are still several limitations: 1) speculative decoding needs great effort to find the right draft model with high level of alignment to ensure acceptance rate, 2) the ratio of the two models needs to break even the efficiency gain, and 3), most crucially, all of these methods require full forward passes of the main model for verification. \ours{} can be seen as a self-speculative model that blindly trusts all drafted tokens with sub-linear verification cost. Additionally, the drafting overhead is constant w.r.t. the generation length. This reduces the complexity of serving, removes alignment process, and saves verification cost, which makes it one of the few speculative-sampling-like solutions for accelerating high-throughput regimes and/or with a high demand for smaller models. 

\subsection{KV-cache Compression}

Key-Value (KV) cache plays a pivotal role in inference generation for many reasons: 1) KV cache grows linearly with the generation length and makes long generation increasingly costly and 2) loading the cache every forward can easily make the decoding phase memory-bounded, hence under-utilizing the full potential of hardware accelerators. To solve these problems, prior works take two major routes~\cite{zhang2023h2oheavyhitteroracleefficient, chen2024magicpiglshsamplingefficient, hooper2024kvquant10millioncontext, xiao2024efficientstreaminglanguagemodels}, namely compressing the cache and retrieve only a subset of them for decoding. SnapKV~\cite{li2024snapkvllmknowslooking} explores special patterns for each head and hence removes unnecessary information. ShadowKV~\cite{sun2025shadowkvkvcacheshadows} employs special low-ranking strategy and CPU offloading for high-throughput optimization. Speculative Prefill~\cite{liu2025speculativeprefillturbochargingttft} and GemFilter~\cite{shi2024discoveringgemsearlylayers} explore sparsity in prompt KVs, which reduces the size consistently for all generated tokens. However, none of these methods \textit{skip computing these KV caches for generated KVs in the first place}. \ours{}, on the other hand, will only compute a single KV for potentially more than on tokens, which is orthogonal to all the compression and offloading methods described above. We want to highlight an important distinction: our method revisits what information a single KV cache can carry, thus breaking the constraints defined by the tokenizer itself. 

\section{Method}

\begin{figure}
    \centering
    \vspace{-0.5cm}
    \begin{subfigure}[b]{\textwidth}
        \includegraphics[width=\textwidth]{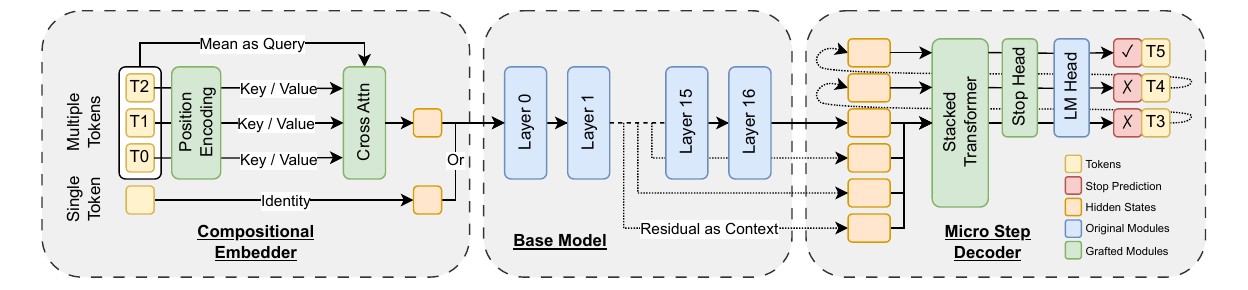}
        \caption{\textbf{Architecture}: \ours{} stacks the base model (middle) with two additional modules. The \textit{light-weight} relative-position-aware compositional embedder (left) fuses a list of token(s) from the previous macro-step: it adopts a cross-attention to create a single hidden state (or bypass if only a single token is given) for the current macro-step forward. The micro-step decoder (right) conditions on several middle layers' hidden states (including the last) as the context to generate micro-step tokens and stop conditions auto-regressively.}
        \label{fig:arch}
    \end{subfigure}
    \begin{subfigure}[b]{\textwidth}
        \includegraphics[width=\textwidth]{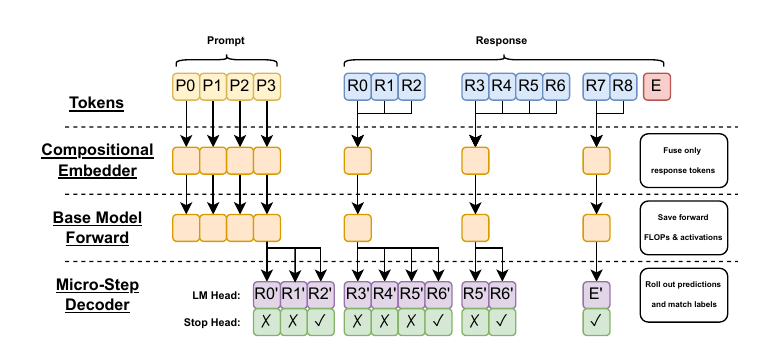}
        \caption{\textbf{Training}: \ours{} trains with standard language modeling objective augmented with an auxiliary binary classification loss: 1) we only calculate loss on response tokens 2) we only fuse tokens belonging to the response 3) the activation and FLOPs of the base model are significantly reduced due to tokens fusions 4) we roll out both token predictions from the language head and stop predictions from the stop head, which are supervised with the ground truth labels of the next macro-step. Here $P_i$ are the prompt tokens, $R_i$ are segmented response tokens, $E$ is the EOS token, letters with prime marks are predicted tokens, and stop predictions are either \ding{55} or \ding{51}.}
        \label{fig:train}
    \end{subfigure}
    \caption{\textbf{\ours{} Overview}: \ours{} stacks the base model with two additional modules that can fuse and predict multiple tokens per iteration for faster decoding. }
\end{figure}

\subsection{\ours{} Architecture}

\ours{} builds on top of the existing popular transformer architecture by stacking a \textit{Compositional Embedder} before the first layer and a \textit{Micro-Step Decoder} between the last layer and the language modeling head, hence the name \ours{} (see Figure~\ref{fig:arch}). The inference procedure switches between two different phrases: 1) a macro-step is defined as a single forward of the base model, followed by 2) several micro-steps where a token is generated from the decoder with a predicted stop condition that decides if we want to move to the next macro-step or perform further micro-step decoding. 

The \textit{Compositional Embedder} consumes an ordered list of token embeddings from the previous macro-step and converts it to a single \textit{permutation-variant} embedding of the same dimension as the original model. This allows computing only a single forward pass and storing one set of KV cache for multiple tokens. The \textit{Micro-Step Decoder} decoder will then take the information from the base model and auto-regressively generate micro-step tokens with a stop condition for each micro-step iteration. We will discuss the two modules as well as the data preparation in more detail next. 

\subsection{Compositional Embedder}
The objective of the \textit{Compositional Embedder} is to learn a mapping from a list of token embeddings to a single hidden state. The first crucially important property is to ensure that tokens can interact in an expressive way while remaining computationally cheap and batch-friendly. To this end, we use a cross-attention mechanism where the mean of the inputs is used as the query to extract information from the list of token embeddings whose keys and values are computed, as inspired by the Perceiver architecture~\cite{jaegle2021perceivergeneralperceptioniterative, pagnoni2024bytelatenttransformerpatches}. In addition, we added learned positional embedding to ensure that relative positional information within the patch is preserved during the fusion process, which we empirically verified to significantly improve tasks that require more fine-grained understanding of math and code. Inspired by Speculative Prefill~\cite{liu2025speculativeprefillturbochargingttft}, we also offset the position information when we send the merged embedding to the base model so that it can be aware of the number of fused tokens from last step. 

\subsection{Micro-Step Decoder}
We use several transformer layers with the same configuration as the base model for our \textit{Micro-Step Decoder}. Since more than one token will potentially share a single KV during the fusion process and hence lose detailed information, we further conditioned our module on several hidden outputs from middle layers with different resolution as the context. We use a separate head that takes in the hidden states and predicts whether we should stop generating micro-step tokens, which is more efficiently, as we experimented, than letting the decoder spend one more micro-step to predict a special stop token. 

During inference, we can optionally also set a threshold for the stop condition so that \ours{} only continues to generate micro-step token if it is confident enough. Although it can potentially decrease the efficiency gain, we found it to be extremely useful for certain tasks where errors introduced by the stop prediction can harm the quality. 

It is worth noting that although we use multiple transformer layers here, the computational cost is constant w.r.t. the context length, which is ideally suitable for long context tasks. The reason is that we have a fixed context consisting of hidden features from the middle layers and a maximum decoding step as defined by the data processing next. 

\subsection{Implicit Regularization for Fast Adaptation and Efficient Inference}

\ours{} essentially creates a dynamic vocabulary whose granularity (i.e. virtual tokens) is defined by what a single KV cache can represent. The space of this new vocabulary will overlap with the original space, and hence making sure they are close enough is the key to successful adaptation. To this end, we revert back to the original computation when the input token list from the last macro-step contains a single token and/or when we predict the first micro-step token in the current macro-step. Specifically, in this case, we skip the \textit{Compositional Embedder} and/or the \textit{Micro-Step Decoder}. The goal here is to implicitly regularize the outputs of our embedder and decoder to match the original embeddings in distribution, and to remove unnecessary overhead during inference. 

\subsection{Dynamic Data Segmentation}
In order to teach the model how many micro-step tokens to output per macro-step, we rely on preprocessing the response of the training SFT data into continuous and non-overlapping segments. Instead of using a fixed window size, we opted for using the knowledge from the base model to determine the capacity of its KV cache. Intuitively, we want to find conditionally confident tokens that can be predicted accurately with local information, so that we don not need to waste full forward passes on them. 

Concretely, given a sequence of input and output token ids $\{x_i\}_N$ and $\{y_i\}_M$, we send it directly to the base model to get the conditional entropy of the output tokens $\_, \{e_i\}_M := f_\theta(\{x_i\}_N, \{y_i\}_M)$. We apply a simple yet effective heuristics based on global statistics to segment the data so that each segment will have the first token with highest entropy above a predefined threshold and the rest with low entropies relative to that of the first token as in Figure~\ref{fig:gen}. 

\subsection{Training}

We conduct supervised fine-tuning (SFT) on top of trained instruct models (training illustration in Figure~\ref{fig:train}). Since we use data with segmented outputs, we will fully expand the micro-step during training (i.e. generate a \texttt{MAX\_STEPS} number of tokens each time) and trim off loss-irrelevant terms. \ours{} only spends one forward for a group of merged tokens, which will drastically reduce the memory requirement due to smaller activations on all modules of the base model and shorter context length for self-attention. 

\subsection{Cost Analysis}

The main efficiency benefits of \ours{} stem from the following features: 
\paragraph{Token Fusion:} Multiple tokens can be fused into a single \textit{virtual} token that occupies a single set of KV cache for the whole forward, which makes long context generation sub-linear. The overall achievable speedup per macro-step is $r:=\frac{n \cdot C(S)}{C(S) + (n - 1) \cdot c} > 1, c < C(S)$ where $n$ is the number of micro-step tokens, $C(S)$ is the cost of the base model that depends on the context length $S$, and $c$ is a constant denoting the cost of the micro-step decoder. 
\paragraph{Minimal Overhead:} When either the input token from the last macro-step is a single token or the current macro-step outputs a single micro-step output token, \ours{} will revert back to the same computation of the base model, resulting in almost zero overhead. 
\paragraph{Batch-friendly and Parallelizable:} Unlike speculative decoding, \ours{} is very batch-friendly because there is no wasted FLOPs for verification and both grafted modules are parallelizable, which makes it suitable for high-throughput regime. 
\paragraph{Practicability with Small Models:} Our method works well even for small models compared to speculative decoding whose expected speedup is $r:=\frac{1 - \alpha^{n+1}}{(1 - \alpha)\cdot(n \cdot c + 1)}$, where $\alpha < 1$ is the acceptance rate and $c$ is the latency ratio of the draft and base model. 
\section{Experiments}

\subsection{Setup}

We build on top of \textsc{Llama-3.2-1B-Instruct} with a \textit{Micro-step Decoder} consisting of four transformer layers of the same configuration. The \textit{Compositional Embedder} is a multi-head cross-attention module followed by a linear layer. 

We train \ours{} for just one epoch with 8 $\times$ H100 Nvidia GPUs using the AdamW optimizer~\cite{loshchilov2019decoupledweightdecayregularization}. Following standard practice, we set the initial learning rates for the base weights to 5\text{e-}5 and those of the randomly initialized grafted weights to 1\text{e-}4. We use a cosine learning rate scheduling with minimum learning rate set to one tenth of the starting value. We choose a global batch size of 128 with the DDP degree equal to 8 and the gradient accumulation step equal to 2. Our training data only includes publicly available datasets as described in Appendix Table~\ref{tab:data} with a maximum micro-step size of 4. 

For a fair comparison, we create a server with common API which process task queries from clients of \texttt{lm\_eval\_harness}~\cite{eval-harness}, \texttt{eval\_plus}~\cite{evalplus} and LongBench code~\cite{bai2024longbenchbilingualmultitaskbenchmark}, and implement our model in GPT-Fast\footnote{\url{https://github.com/pytorch-labs/gpt-fast}} for inference benchmarking. All results are generated greedily with zero-shot. 

\subsection{Downstream Performance}

We evaluate \ours{} on both standard tasks in~\ref{sec:standard} and long-context tasks in~\ref{sec:long} for completeness. We vary the confidence level which decides whether we want to continue micro-step generation based on model predicted stop condition, which trade-offs between quality and speedup / KV cache compression rate. All of our chosen tasks are generative to make sure unbiased and fair evaluation. 

\subsubsection{Standard Tasks}
\label{sec:standard}

\begin{figure}[ht]
\centering
\vspace{-0.5cm}
\makebox[\textwidth][c]{\includegraphics[width=0.9\textwidth]{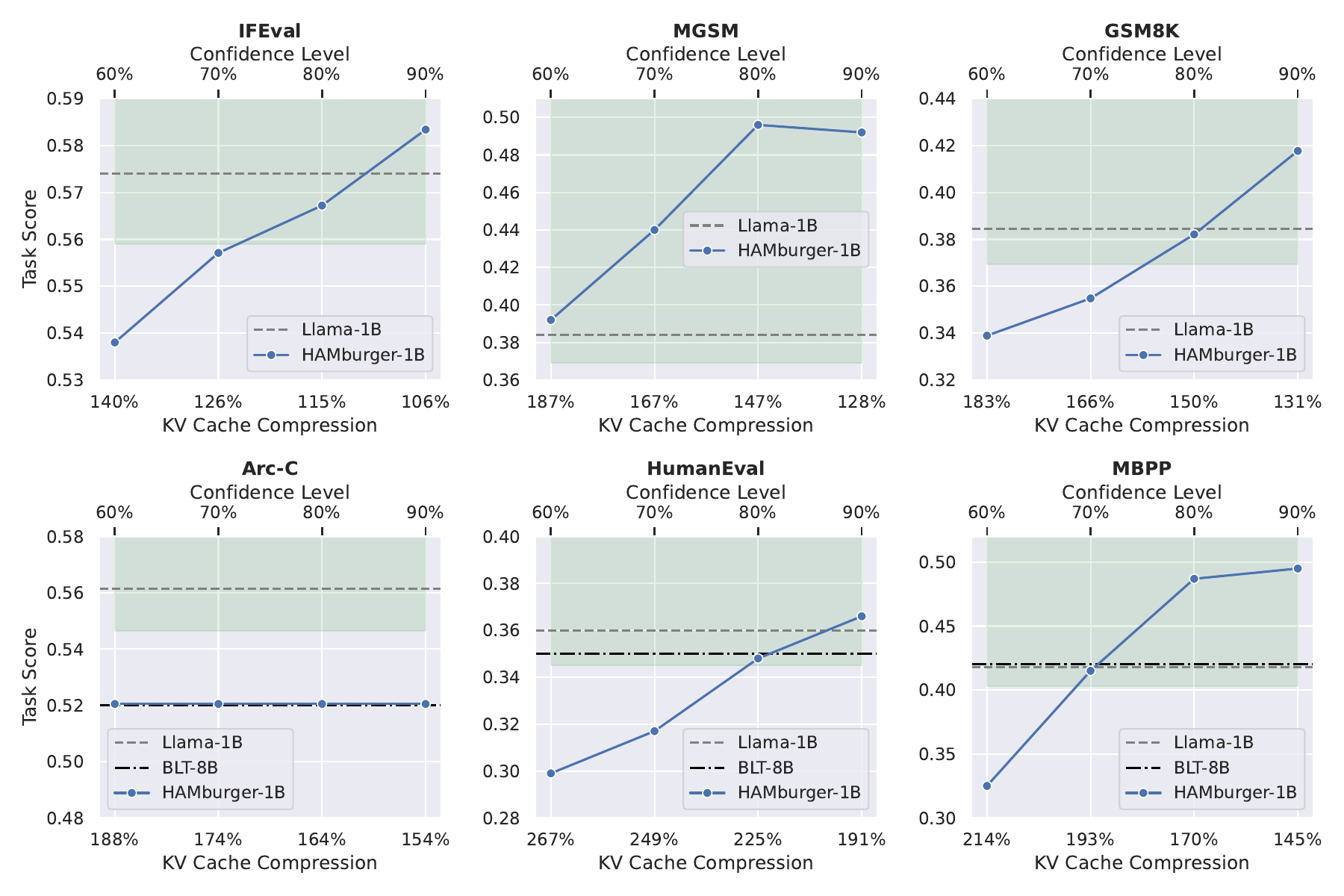}}
\vspace{-0.5cm}
\caption{\textbf{Standard Task Evaluation:} We present the standard task evaluations for our method, which consists of instruction following, math, reasoning and code. The bottom x-axis denotes the KV cache compression rate, which relates to the decoding TPS speedup. The top x-axis shows a tunable parameter that trades-off efficiency and quality. We shade the green area to be the bearable quality loss. In almost all tasks, \ours{} achieves great efficiency with minimal quality loss. }
\label{fig:standard}
\end{figure}

We start by evaluating \ours{} in representative standard tasks. In Figure~\ref{fig:standard}, we selected domains that span instruction following (IFEval~\cite{zhou2023instructionfollowingevaluationlargelanguage}), math (zero-shot GSM8K~\cite{cobbe2021trainingverifierssolvemath}), reasoning(MGSM~\cite{shi2022languagemodelsmultilingualchainofthought} and Arc-C~\cite{clark2018thinksolvedquestionanswering}), and coding ability (HumanEval~\cite{chen2021evaluatinglargelanguagemodels} and MBPP~\cite{austin2021programsynthesislargelanguage}). As we can see, \ours{} can maintain and even surpass baseline and \textsc{BLT-8B} with the right confidence level in terms of quality, and provide substantially reduced KV cache computation and hence inference latency. We found that our method falls a bit short on Arc-C and upon examining the task prompts, we think it is mostly due to the fact that it asks for direct multiple choice answer without intermediate step. In order to show that \ours{} does not have deficient reasoning ability, we compliment Arc-C with MGSM in English which involves chain-of-thought reasoning and has higher requirement for instruction following to produce well-formatted outputs. We demonstrate that \ours{} can achieve competitive quality while performing inference with a much lower cost and latency. 

\subsubsection{Long Context Tasks}
\label{sec:long}

\begin{figure}[h]
\centering
\makebox[\textwidth][c]{\includegraphics[width=1.1\textwidth]{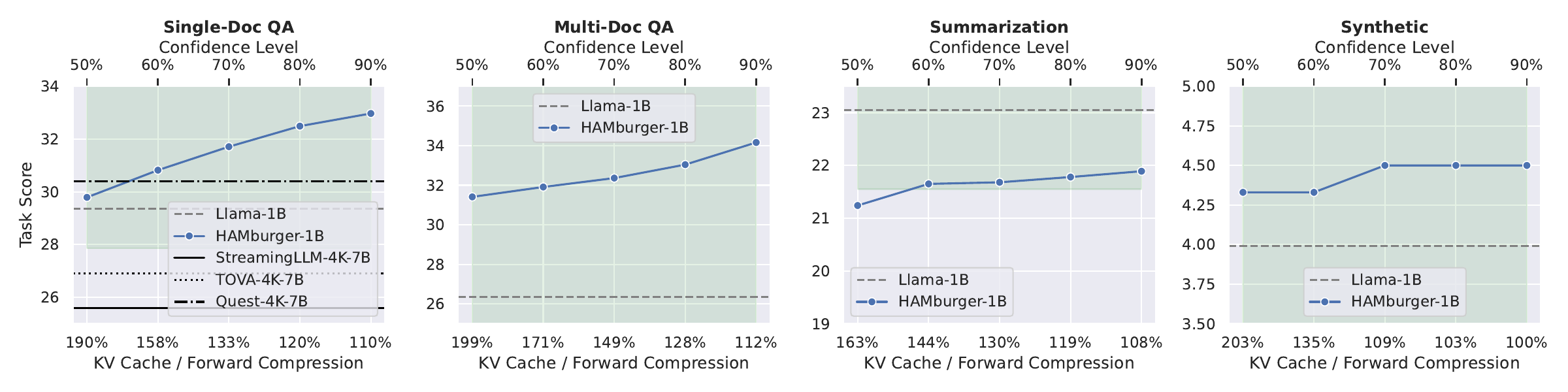}}
\vspace{-0.5cm}
\caption{\textbf{Long Context Task Evaluation:} We showcase the superior long-context performance of \ours{} with LongBench suite. With the same graph settings as Figure~\ref{fig:standard}, we can see that our method performs competitively quality-wise while significantly reducing the serving cost and latency. }
\label{fig:longbench}
\end{figure}

\ours{} significantly optimizes the decoding phase by dynamically generating multiple tokens per step and we want to show that this holds true for long-context applications as well. We choose LongBench~\cite{bai2024longbenchbilingualmultitaskbenchmark} which consists of many important tasks for probing long context capability. Since \ours{} is an instruct model and its dynamic decoding requires recognizing the start of the assistant from the chat template, we focus on four main domains as shown in Figure~\ref{fig:longbench}. For all tasks except summarization, \ours{} outperforms the baseline by margins with up to 2$\times$ reduction in KV cache computation, and hence reduces latency proportionally. To break down summarization, we found that \ours{} falls behind on GovReport~\cite{huang2021efficientattentionslongdocument} only, and is better at the other task MultiNews~\cite{fabbri2019multinewslargescalemultidocumentsummarization}, resulting in slight lower average. Upon examining the training data mix, we believe that this is mostly due to missing similar data samples. We also show that our 1B model with 60\% confidence level outperforms sparse KV methods like StreamingLLM~\cite{xiao2024efficientstreaminglanguagemodels}, TOVA~\cite{oren2024transformersmultistaternns}, and Quest~\cite{tang2024questqueryawaresparsityefficient} with 4096 KV budget on \textsc{LongChat-7b-v1.5-32k}\footnote{They only tested the full sub-tasks for Single-Doc QA.}. Overall, we believe \ours{} is fully long-context capable and we leave finding a better data mix as the future work. 

\subsection{Efficiency Benchmarking}
\label{sec:efficiency}

\begin{figure}[h]
\centering
\makebox[\textwidth][c]{\includegraphics[width=0.8\textwidth]{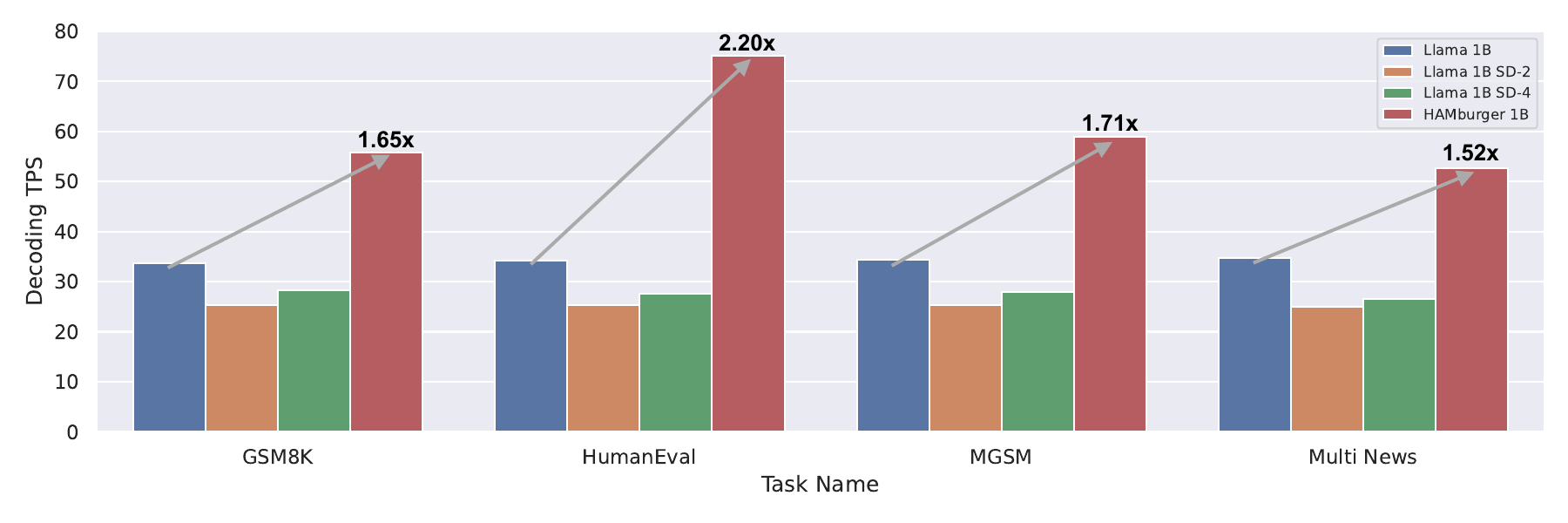}}
\vspace{-0.5cm}
\caption{\textbf{Efficiency Benchmarking:} We compare the decoding tokens per second for \ours{} against the baselines. For 1B models, \ours{} achieves up to 2.2$\times$ decoding TPS speedup over the base model. \ours{} also beats speculative sampling with 1B INT8 draft model with even $>$ 90\% acceptance rate by speedup (up to 2.73$\times$) and memory saving. }
\label{fig:efficiency}
\end{figure}

To fully understand what the real decoding speed improvement \ours{} can achieve, we implement it in GPT-Fast for a balance of efficiency and simplicity. Since the generation dynamics of \ours{} depends on the prompt itself, we use samples from GSM8K~\cite{cobbe2021trainingverifierssolvemath}, HumanEval~\cite{chen2021evaluatinglargelanguagemodels}, MGSM~\cite{shi2022languagemodelsmultilingualchainofthought}, MultiNews~\cite{fabbri2019multinewslargescalemultidocumentsummarization} to benchmark its decoding TPS over different task domains. In Figure~\ref{fig:efficiency}, we compare \ours{} against running the base model and speculative decoding of which our method can be considered as a special case (i.e., \ours{} does self-speculation without rejection). 

First of all, we can observe a decoding TPS speedup from 1.52$\times$ to 2.20$\times$ with proportional memory reduction. Furthermore, we want to show that \ours{} can still provide valuable contribution for smaller models, an extremely challenging regime for speculative decoding. In order to match tokenizer vocabulary and emulate the upper performance bound, we use an INT8 version of \textsc{Llama-3.2-1B-Instruct} as the draft model for speculative decoding, which ends up achieving > 90\% acceptance rate for both $\text{gamma}=2$ and $\text{gamma}=4$. As we can observe, speculative decoding in both settings are eclipsed by \ours{} since finding a high quality while still fast drafter for smaller model is non-trivial. Finally, we want to highlight that the micro-step FLOPs is constant w.r.t. the generation length, making it extremely suitable for long prompts.

\subsection{Ablation}

\begin{table}[h]
\centering
\caption{\textbf{Ablation on Architecture Design}: We present results on three most representative tasks that showcase the benefits of our core architecture designs. }
\label{tab:ablation}
\vspace{0.2cm}
\begin{tabular}{l|ccc|c}
\toprule
\textbf{Architecture Change} & \textbf{GSM8K} & \textbf{Arc-C} & \textbf{HumanEval} & \textbf{Eval > 1 Token Acc} \\
\midrule
\midrule
Base & 0.18 & 0.49 & 0.11 & 0.55 \\
\quad\textit{w/ Middle Hidden Features} & 0.20 & 0.50 & 0.12 & 0.76\\
\quad\textit{w/ Cross Attn Embedder} & 0.33  & 0.55 & 0.35 & 0.92 \\
\quad\textit{w/ Stop Prediction (final)} & 0.38  & 0.56 & 0.33 & 0.93 \\

\bottomrule
\end{tabular}
\end{table}

In this section, we empirically prove the effectiveness of our core architecture designs in terms of model performance. In Table~\ref{tab:ablation}, we choose three representative downstream tasks and the average token accuracy (except for the first micro-token) of the evaluation set for ablation. 

First, adding middle layers' hidden features as the context will significantly improve the modeling ability of our \textit{Micro-step Decoder} as shown by the average evaluation token accuracy excluding the first micro-step token. This meets our expectation since the first micro-token already received sufficient information from the last hidden states and adding the context helps better situate micro-step predictions starting at the second one. 

After that, replacing a simple per-dimension learned Softmax merging with our cross-attention based \textit{Compositional embedder} significantly improves the quality for tasks that require finer-grained understanding of token details (e.g., GSM8K~\cite{cobbe2021trainingverifierssolvemath}, Arc Challenge~\cite{clark2018thinksolvedquestionanswering} and HumanEval~\cite{chen2021evaluatinglargelanguagemodels}), since it can capture more nuanced and higher-order interactions among the input list of tokens. 

Finally, we switched from stopping micro-token generation based on predicting a special reserved token from the vocabulary using one more micro-step, to using a separate stop head for binary classification. We found that this change maintains similar task performance but provides remarkable efficiency improvements due to reduced overhead (i.e., we do not need to pay one extra micro-step forward every macro-step). 

\section{Limitation and Future Work}

Although with current instruction fine-tuning and the curated data mix, our model can already perform well, we believe future research on designing post-generation-aware segmentation strategy can be helpful, especially when the compositional embedder is not perfect. We also believe \ours{} belongs to very few works that can make speculators themselves faster, but deciding how to optimally handle newly added token from the base model requires more careful consideration. 

\section{Conclusion}

In this work, we present \ours{}, which redefines resource allocation in LLMs by moving beyond uniform computation and storage per token during inference. With the help of a relative-position-aware compositional embedder for token fusion and a local micro-step decoder that outputs multiple tokens per step, our method achieves up to 2$\times$ reduction in KV cache \textit{computation} (hence storage) and up to 2 $\times$ TPS improvement while maintaining or even surpassing the base model in quality. Given reduced the linear growth rate of KV cache and forward FLOPs to sub-linear w.r.t. the output length, we believe that \ours{} will provide valuable inspiration and insights to future LLM architectures that are information-conscious and dynamic in nature.

\bibliographystyle{abbrv}
\bibliography{main}

\begin{thebibliography}{10}

\bibitem{claude2025}
Anthropic.
\newblock The claude 3 model family: Opus, sonnet, haiku.
\newblock \url{https://www-cdn.anthropic.com/de8ba9b01c9ab7cbabf5c33b80b7bbc618857627/Model_Card_Claude_3.pdf}, 2025.

\bibitem{austin2021programsynthesislargelanguage}
J.~Austin, A.~Odena, M.~Nye, M.~Bosma, H.~Michalewski, D.~Dohan, E.~Jiang, C.~Cai, M.~Terry, Q.~Le, and C.~Sutton.
\newblock Program synthesis with large language models, 2021.

\bibitem{bai2024longbenchbilingualmultitaskbenchmark}
Y.~Bai, X.~Lv, J.~Zhang, H.~Lyu, J.~Tang, Z.~Huang, Z.~Du, X.~Liu, A.~Zeng, L.~Hou, Y.~Dong, J.~Tang, and J.~Li.
\newblock Longbench: A bilingual, multitask benchmark for long context understanding, 2024.

\bibitem{cai2024medusasimplellminference}
T.~Cai, Y.~Li, Z.~Geng, H.~Peng, J.~D. Lee, D.~Chen, and T.~Dao.
\newblock Medusa: Simple llm inference acceleration framework with multiple decoding heads, 2024.

\bibitem{chen2023acceleratinglargelanguagemodel}
C.~Chen, S.~Borgeaud, G.~Irving, J.-B. Lespiau, L.~Sifre, and J.~Jumper.
\newblock Accelerating large language model decoding with speculative sampling, 2023.

\bibitem{chen2021evaluatinglargelanguagemodels}
M.~Chen, J.~Tworek, H.~Jun, Q.~Yuan, H.~P. de~Oliveira~Pinto, J.~Kaplan, H.~Edwards, Y.~Burda, N.~Joseph, G.~Brockman, A.~Ray, R.~Puri, G.~Krueger, M.~Petrov, H.~Khlaaf, G.~Sastry, P.~Mishkin, B.~Chan, S.~Gray, N.~Ryder, M.~Pavlov, A.~Power, L.~Kaiser, M.~Bavarian, C.~Winter, P.~Tillet, F.~P. Such, D.~Cummings, M.~Plappert, F.~Chantzis, E.~Barnes, A.~Herbert-Voss, W.~H. Guss, A.~Nichol, A.~Paino, N.~Tezak, J.~Tang, I.~Babuschkin, S.~Balaji, S.~Jain, W.~Saunders, C.~Hesse, A.~N. Carr, J.~Leike, J.~Achiam, V.~Misra, E.~Morikawa, A.~Radford, M.~Knight, M.~Brundage, M.~Murati, K.~Mayer, P.~Welinder, B.~McGrew, D.~Amodei, S.~McCandlish, I.~Sutskever, and W.~Zaremba.
\newblock Evaluating large language models trained on code, 2021.

\bibitem{chen2024magicpiglshsamplingefficient}
Z.~Chen, R.~Sadhukhan, Z.~Ye, Y.~Zhou, J.~Zhang, N.~Nolte, Y.~Tian, M.~Douze, L.~Bottou, Z.~Jia, and B.~Chen.
\newblock Magicpig: Lsh sampling for efficient llm generation, 2024.

\bibitem{clark2018thinksolvedquestionanswering}
P.~Clark, I.~Cowhey, O.~Etzioni, T.~Khot, A.~Sabharwal, C.~Schoenick, and O.~Tafjord.
\newblock Think you have solved question answering? try arc, the ai2 reasoning challenge, 2018.

\bibitem{cobbe2021trainingverifierssolvemath}
K.~Cobbe, V.~Kosaraju, M.~Bavarian, M.~Chen, H.~Jun, L.~Kaiser, M.~Plappert, J.~Tworek, J.~Hilton, R.~Nakano, C.~Hesse, and J.~Schulman.
\newblock Training verifiers to solve math word problems, 2021.

\bibitem{deepseekai2025deepseekv3technicalreport}
DeepSeek-AI, A.~Liu, B.~Feng, B.~Xue, B.~Wang, B.~Wu, C.~Lu, C.~Zhao, C.~Deng, C.~Zhang, C.~Ruan, D.~Dai, D.~Guo, D.~Yang, D.~Chen, D.~Ji, E.~Li, F.~Lin, F.~Dai, F.~Luo, G.~Hao, G.~Chen, G.~Li, H.~Zhang, H.~Bao, H.~Xu, H.~Wang, H.~Zhang, H.~Ding, H.~Xin, H.~Gao, H.~Li, H.~Qu, J.~L. Cai, J.~Liang, J.~Guo, J.~Ni, J.~Li, J.~Wang, J.~Chen, J.~Chen, J.~Yuan, J.~Qiu, J.~Li, J.~Song, K.~Dong, K.~Hu, K.~Gao, K.~Guan, K.~Huang, K.~Yu, L.~Wang, L.~Zhang, L.~Xu, L.~Xia, L.~Zhao, L.~Wang, L.~Zhang, M.~Li, M.~Wang, M.~Zhang, M.~Zhang, M.~Tang, M.~Li, N.~Tian, P.~Huang, P.~Wang, P.~Zhang, Q.~Wang, Q.~Zhu, Q.~Chen, Q.~Du, R.~J. Chen, R.~L. Jin, R.~Ge, R.~Zhang, R.~Pan, R.~Wang, R.~Xu, R.~Zhang, R.~Chen, S.~S. Li, S.~Lu, S.~Zhou, S.~Chen, S.~Wu, S.~Ye, S.~Ye, S.~Ma, S.~Wang, S.~Zhou, S.~Yu, S.~Zhou, S.~Pan, T.~Wang, T.~Yun, T.~Pei, T.~Sun, W.~L. Xiao, W.~Zeng, W.~Zhao, W.~An, W.~Liu, W.~Liang, W.~Gao, W.~Yu, W.~Zhang, X.~Q. Li, X.~Jin, X.~Wang, X.~Bi, X.~Liu, X.~Wang, X.~Shen, X.~Chen, X.~Zhang, X.~Chen, X.~Nie, X.~Sun,
  X.~Wang, X.~Cheng, X.~Liu, X.~Xie, X.~Liu, X.~Yu, X.~Song, X.~Shan, X.~Zhou, X.~Yang, X.~Li, X.~Su, X.~Lin, Y.~K. Li, Y.~Q. Wang, Y.~X. Wei, Y.~X. Zhu, Y.~Zhang, Y.~Xu, Y.~Xu, Y.~Huang, Y.~Li, Y.~Zhao, Y.~Sun, Y.~Li, Y.~Wang, Y.~Yu, Y.~Zheng, Y.~Zhang, Y.~Shi, Y.~Xiong, Y.~He, Y.~Tang, Y.~Piao, Y.~Wang, Y.~Tan, Y.~Ma, Y.~Liu, Y.~Guo, Y.~Wu, Y.~Ou, Y.~Zhu, Y.~Wang, Y.~Gong, Y.~Zou, Y.~He, Y.~Zha, Y.~Xiong, Y.~Ma, Y.~Yan, Y.~Luo, Y.~You, Y.~Liu, Y.~Zhou, Z.~F. Wu, Z.~Z. Ren, Z.~Ren, Z.~Sha, Z.~Fu, Z.~Xu, Z.~Huang, Z.~Zhang, Z.~Xie, Z.~Zhang, Z.~Hao, Z.~Gou, Z.~Ma, Z.~Yan, Z.~Shao, Z.~Xu, Z.~Wu, Z.~Zhang, Z.~Li, Z.~Gu, Z.~Zhu, Z.~Liu, Z.~Li, Z.~Xie, Z.~Song, Z.~Gao, and Z.~Pan.
\newblock Deepseek-v3 technical report, 2025.

\bibitem{fabbri2019multinewslargescalemultidocumentsummarization}
A.~R. Fabbri, I.~Li, T.~She, S.~Li, and D.~R. Radev.
\newblock Multi-news: a large-scale multi-document summarization dataset and abstractive hierarchical model, 2019.

\bibitem{feng2024faragillmsneed}
T.~Feng, C.~Jin, J.~Liu, K.~Zhu, H.~Tu, Z.~Cheng, G.~Lin, and J.~You.
\newblock How far are we from agi: Are llms all we need?, 2024.

\bibitem{eval-harness}
L.~Gao, J.~Tow, B.~Abbasi, S.~Biderman, S.~Black, A.~DiPofi, C.~Foster, L.~Golding, J.~Hsu, A.~Le~Noac'h, H.~Li, K.~McDonell, N.~Muennighoff, C.~Ociepa, J.~Phang, L.~Reynolds, H.~Schoelkopf, A.~Skowron, L.~Sutawika, E.~Tang, A.~Thite, B.~Wang, K.~Wang, and A.~Zou.
\newblock The language model evaluation harness, 07 2024.

\bibitem{grattafiori2024llama3herdmodels}
A.~Grattafiori, A.~Dubey, A.~Jauhri, A.~Pandey, A.~Kadian, A.~Al-Dahle, A.~Letman, A.~Mathur, A.~Schelten, A.~Vaughan, A.~Yang, A.~Fan, A.~Goyal, A.~Hartshorn, A.~Yang, A.~Mitra, A.~Sravankumar, A.~Korenev, A.~Hinsvark, A.~Rao, A.~Zhang, A.~Rodriguez, A.~Gregerson, A.~Spataru, B.~Roziere, B.~Biron, B.~Tang, B.~Chern, C.~Caucheteux, C.~Nayak, C.~Bi, C.~Marra, C.~McConnell, C.~Keller, C.~Touret, C.~Wu, C.~Wong, C.~C. Ferrer, C.~Nikolaidis, D.~Allonsius, D.~Song, D.~Pintz, D.~Livshits, D.~Wyatt, D.~Esiobu, D.~Choudhary, D.~Mahajan, D.~Garcia-Olano, D.~Perino, D.~Hupkes, E.~Lakomkin, E.~AlBadawy, E.~Lobanova, E.~Dinan, E.~M. Smith, F.~Radenovic, F.~Guzmán, F.~Zhang, G.~Synnaeve, G.~Lee, G.~L. Anderson, G.~Thattai, G.~Nail, G.~Mialon, G.~Pang, G.~Cucurell, H.~Nguyen, H.~Korevaar, H.~Xu, H.~Touvron, I.~Zarov, I.~A. Ibarra, I.~Kloumann, I.~Misra, I.~Evtimov, J.~Zhang, J.~Copet, J.~Lee, J.~Geffert, J.~Vranes, J.~Park, J.~Mahadeokar, J.~Shah, J.~van~der Linde, J.~Billock, J.~Hong, J.~Lee, J.~Fu, J.~Chi, J.~Huang,
  J.~Liu, J.~Wang, J.~Yu, J.~Bitton, J.~Spisak, J.~Park, J.~Rocca, J.~Johnstun, J.~Saxe, J.~Jia, K.~V. Alwala, K.~Prasad, K.~Upasani, K.~Plawiak, K.~Li, K.~Heafield, K.~Stone, K.~El-Arini, K.~Iyer, K.~Malik, K.~Chiu, K.~Bhalla, K.~Lakhotia, L.~Rantala-Yeary, L.~van~der Maaten, L.~Chen, L.~Tan, L.~Jenkins, L.~Martin, L.~Madaan, L.~Malo, L.~Blecher, L.~Landzaat, L.~de~Oliveira, M.~Muzzi, M.~Pasupuleti, M.~Singh, M.~Paluri, M.~Kardas, M.~Tsimpoukelli, M.~Oldham, M.~Rita, M.~Pavlova, M.~Kambadur, M.~Lewis, M.~Si, M.~K. Singh, M.~Hassan, N.~Goyal, N.~Torabi, N.~Bashlykov, N.~Bogoychev, N.~Chatterji, N.~Zhang, O.~Duchenne, O.~Çelebi, P.~Alrassy, P.~Zhang, P.~Li, P.~Vasic, P.~Weng, P.~Bhargava, P.~Dubal, P.~Krishnan, P.~S. Koura, P.~Xu, Q.~He, Q.~Dong, R.~Srinivasan, R.~Ganapathy, R.~Calderer, R.~S. Cabral, R.~Stojnic, R.~Raileanu, R.~Maheswari, R.~Girdhar, R.~Patel, R.~Sauvestre, R.~Polidoro, R.~Sumbaly, R.~Taylor, R.~Silva, R.~Hou, R.~Wang, S.~Hosseini, S.~Chennabasappa, S.~Singh, S.~Bell, S.~S. Kim, S.~Edunov,
  S.~Nie, S.~Narang, S.~Raparthy, S.~Shen, S.~Wan, S.~Bhosale, S.~Zhang, S.~Vandenhende, S.~Batra, S.~Whitman, S.~Sootla, S.~Collot, S.~Gururangan, S.~Borodinsky, T.~Herman, T.~Fowler, T.~Sheasha, T.~Georgiou, T.~Scialom, T.~Speckbacher, T.~Mihaylov, T.~Xiao, U.~Karn, V.~Goswami, V.~Gupta, V.~Ramanathan, V.~Kerkez, V.~Gonguet, V.~Do, V.~Vogeti, V.~Albiero, V.~Petrovic, W.~Chu, W.~Xiong, W.~Fu, W.~Meers, X.~Martinet, X.~Wang, X.~Wang, X.~E. Tan, X.~Xia, X.~Xie, X.~Jia, X.~Wang, Y.~Goldschlag, Y.~Gaur, Y.~Babaei, Y.~Wen, Y.~Song, Y.~Zhang, Y.~Li, Y.~Mao, Z.~D. Coudert, Z.~Yan, Z.~Chen, Z.~Papakipos, A.~Singh, A.~Srivastava, A.~Jain, A.~Kelsey, A.~Shajnfeld, A.~Gangidi, A.~Victoria, A.~Goldstand, A.~Menon, A.~Sharma, A.~Boesenberg, A.~Baevski, A.~Feinstein, A.~Kallet, A.~Sangani, A.~Teo, A.~Yunus, A.~Lupu, A.~Alvarado, A.~Caples, A.~Gu, A.~Ho, A.~Poulton, A.~Ryan, A.~Ramchandani, A.~Dong, A.~Franco, A.~Goyal, A.~Saraf, A.~Chowdhury, A.~Gabriel, A.~Bharambe, A.~Eisenman, A.~Yazdan, B.~James, B.~Maurer,
  B.~Leonhardi, B.~Huang, B.~Loyd, B.~D. Paola, B.~Paranjape, B.~Liu, B.~Wu, B.~Ni, B.~Hancock, B.~Wasti, B.~Spence, B.~Stojkovic, B.~Gamido, B.~Montalvo, C.~Parker, C.~Burton, C.~Mejia, C.~Liu, C.~Wang, C.~Kim, C.~Zhou, C.~Hu, C.-H. Chu, C.~Cai, C.~Tindal, C.~Feichtenhofer, C.~Gao, D.~Civin, D.~Beaty, D.~Kreymer, D.~Li, D.~Adkins, D.~Xu, D.~Testuggine, D.~David, D.~Parikh, D.~Liskovich, D.~Foss, D.~Wang, D.~Le, D.~Holland, E.~Dowling, E.~Jamil, E.~Montgomery, E.~Presani, E.~Hahn, E.~Wood, E.-T. Le, E.~Brinkman, E.~Arcaute, E.~Dunbar, E.~Smothers, F.~Sun, F.~Kreuk, F.~Tian, F.~Kokkinos, F.~Ozgenel, F.~Caggioni, F.~Kanayet, F.~Seide, G.~M. Florez, G.~Schwarz, G.~Badeer, G.~Swee, G.~Halpern, G.~Herman, G.~Sizov, Guangyi, Zhang, G.~Lakshminarayanan, H.~Inan, H.~Shojanazeri, H.~Zou, H.~Wang, H.~Zha, H.~Habeeb, H.~Rudolph, H.~Suk, H.~Aspegren, H.~Goldman, H.~Zhan, I.~Damlaj, I.~Molybog, I.~Tufanov, I.~Leontiadis, I.-E. Veliche, I.~Gat, J.~Weissman, J.~Geboski, J.~Kohli, J.~Lam, J.~Asher, J.-B. Gaya, J.~Marcus,
  J.~Tang, J.~Chan, J.~Zhen, J.~Reizenstein, J.~Teboul, J.~Zhong, J.~Jin, J.~Yang, J.~Cummings, J.~Carvill, J.~Shepard, J.~McPhie, J.~Torres, J.~Ginsburg, J.~Wang, K.~Wu, K.~H. U, K.~Saxena, K.~Khandelwal, K.~Zand, K.~Matosich, K.~Veeraraghavan, K.~Michelena, K.~Li, K.~Jagadeesh, K.~Huang, K.~Chawla, K.~Huang, L.~Chen, L.~Garg, L.~A, L.~Silva, L.~Bell, L.~Zhang, L.~Guo, L.~Yu, L.~Moshkovich, L.~Wehrstedt, M.~Khabsa, M.~Avalani, M.~Bhatt, M.~Mankus, M.~Hasson, M.~Lennie, M.~Reso, M.~Groshev, M.~Naumov, M.~Lathi, M.~Keneally, M.~Liu, M.~L. Seltzer, M.~Valko, M.~Restrepo, M.~Patel, M.~Vyatskov, M.~Samvelyan, M.~Clark, M.~Macey, M.~Wang, M.~J. Hermoso, M.~Metanat, M.~Rastegari, M.~Bansal, N.~Santhanam, N.~Parks, N.~White, N.~Bawa, N.~Singhal, N.~Egebo, N.~Usunier, N.~Mehta, N.~P. Laptev, N.~Dong, N.~Cheng, O.~Chernoguz, O.~Hart, O.~Salpekar, O.~Kalinli, P.~Kent, P.~Parekh, P.~Saab, P.~Balaji, P.~Rittner, P.~Bontrager, P.~Roux, P.~Dollar, P.~Zvyagina, P.~Ratanchandani, P.~Yuvraj, Q.~Liang, R.~Alao, R.~Rodriguez,
  R.~Ayub, R.~Murthy, R.~Nayani, R.~Mitra, R.~Parthasarathy, R.~Li, R.~Hogan, R.~Battey, R.~Wang, R.~Howes, R.~Rinott, S.~Mehta, S.~Siby, S.~J. Bondu, S.~Datta, S.~Chugh, S.~Hunt, S.~Dhillon, S.~Sidorov, S.~Pan, S.~Mahajan, S.~Verma, S.~Yamamoto, S.~Ramaswamy, S.~Lindsay, S.~Lindsay, S.~Feng, S.~Lin, S.~C. Zha, S.~Patil, S.~Shankar, S.~Zhang, S.~Zhang, S.~Wang, S.~Agarwal, S.~Sajuyigbe, S.~Chintala, S.~Max, S.~Chen, S.~Kehoe, S.~Satterfield, S.~Govindaprasad, S.~Gupta, S.~Deng, S.~Cho, S.~Virk, S.~Subramanian, S.~Choudhury, S.~Goldman, T.~Remez, T.~Glaser, T.~Best, T.~Koehler, T.~Robinson, T.~Li, T.~Zhang, T.~Matthews, T.~Chou, T.~Shaked, V.~Vontimitta, V.~Ajayi, V.~Montanez, V.~Mohan, V.~S. Kumar, V.~Mangla, V.~Ionescu, V.~Poenaru, V.~T. Mihailescu, V.~Ivanov, W.~Li, W.~Wang, W.~Jiang, W.~Bouaziz, W.~Constable, X.~Tang, X.~Wu, X.~Wang, X.~Wu, X.~Gao, Y.~Kleinman, Y.~Chen, Y.~Hu, Y.~Jia, Y.~Qi, Y.~Li, Y.~Zhang, Y.~Zhang, Y.~Adi, Y.~Nam, Yu, Wang, Y.~Zhao, Y.~Hao, Y.~Qian, Y.~Li, Y.~He, Z.~Rait, Z.~DeVito,
  Z.~Rosnbrick, Z.~Wen, Z.~Yang, Z.~Zhao, and Z.~Ma.
\newblock The llama 3 herd of models, 2024.

\bibitem{ho2024block}
N.~Ho, S.~Bae, T.~Kim, H.~Jo, Y.~Kim, T.~Schuster, A.~Fisch, J.~Thorne, and S.-Y. Yun.
\newblock Block transformer: Global-to-local language modeling for fast inference.
\newblock {\em arXiv preprint arXiv:2406.02657}, 2024.

\bibitem{hooper2024kvquant10millioncontext}
C.~Hooper, S.~Kim, H.~Mohammadzadeh, M.~W. Mahoney, Y.~S. Shao, K.~Keutzer, and A.~Gholami.
\newblock Kvquant: Towards 10 million context length llm inference with kv cache quantization, 2024.

\bibitem{huang2021efficientattentionslongdocument}
L.~Huang, S.~Cao, N.~Parulian, H.~Ji, and L.~Wang.
\newblock Efficient attentions for long document summarization, 2021.

\bibitem{jaegle2021perceivergeneralperceptioniterative}
A.~Jaegle, F.~Gimeno, A.~Brock, A.~Zisserman, O.~Vinyals, and J.~Carreira.
\newblock Perceiver: General perception with iterative attention, 2021.

\bibitem{leviathan2023fastinferencetransformersspeculative}
Y.~Leviathan, M.~Kalman, and Y.~Matias.
\newblock Fast inference from transformers via speculative decoding, 2023.

\bibitem{li2024snapkvllmknowslooking}
Y.~Li, Y.~Huang, B.~Yang, B.~Venkitesh, A.~Locatelli, H.~Ye, T.~Cai, P.~Lewis, and D.~Chen.
\newblock Snapkv: Llm knows what you are looking for before generation, 2024.

\bibitem{li2024chunkdistilledlanguagemodeling}
Y.~Li, K.~Livescu, and J.~Zhou.
\newblock Chunk-distilled language modeling, 2024.

\bibitem{li2024eagle2fasterinferencelanguage}
Y.~Li, F.~Wei, C.~Zhang, and H.~Zhang.
\newblock Eagle-2: Faster inference of language models with dynamic draft trees, 2024.

\bibitem{li2025eaglespeculativesamplingrequires}
Y.~Li, F.~Wei, C.~Zhang, and H.~Zhang.
\newblock Eagle: Speculative sampling requires rethinking feature uncertainty, 2025.

\bibitem{liang2023holisticevaluationlanguagemodels}
P.~Liang, R.~Bommasani, T.~Lee, D.~Tsipras, D.~Soylu, M.~Yasunaga, Y.~Zhang, D.~Narayanan, Y.~Wu, A.~Kumar, B.~Newman, B.~Yuan, B.~Yan, C.~Zhang, C.~Cosgrove, C.~D. Manning, C.~Ré, D.~Acosta-Navas, D.~A. Hudson, E.~Zelikman, E.~Durmus, F.~Ladhak, F.~Rong, H.~Ren, H.~Yao, J.~Wang, K.~Santhanam, L.~Orr, L.~Zheng, M.~Yuksekgonul, M.~Suzgun, N.~Kim, N.~Guha, N.~Chatterji, O.~Khattab, P.~Henderson, Q.~Huang, R.~Chi, S.~M. Xie, S.~Santurkar, S.~Ganguli, T.~Hashimoto, T.~Icard, T.~Zhang, V.~Chaudhary, W.~Wang, X.~Li, Y.~Mai, Y.~Zhang, and Y.~Koreeda.
\newblock Holistic evaluation of language models, 2023.

\bibitem{liu2025speculativeprefillturbochargingttft}
J.~Liu, B.~Chen, and C.~Zhang.
\newblock Speculative prefill: Turbocharging ttft with lightweight and training-free token importance estimation, 2025.

\bibitem{evalplus}
J.~Liu, C.~S. Xia, Y.~Wang, and L.~Zhang.
\newblock Is your code generated by chat{GPT} really correct? rigorous evaluation of large language models for code generation.
\newblock In {\em Thirty-seventh Conference on Neural Information Processing Systems}, 2023.

\bibitem{loshchilov2019decoupledweightdecayregularization}
I.~Loshchilov and F.~Hutter.
\newblock Decoupled weight decay regularization, 2019.

\bibitem{miao2023efficientgenerativelargelanguage}
X.~Miao, G.~Oliaro, Z.~Zhang, X.~Cheng, H.~Jin, T.~Chen, and Z.~Jia.
\newblock Towards efficient generative large language model serving: A survey from algorithms to systems, 2023.

\bibitem{openai2024gpt4technicalreport}
OpenAI, J.~Achiam, S.~Adler, S.~Agarwal, L.~Ahmad, I.~Akkaya, F.~L. Aleman, D.~Almeida, J.~Altenschmidt, S.~Altman, S.~Anadkat, R.~Avila, I.~Babuschkin, S.~Balaji, V.~Balcom, P.~Baltescu, H.~Bao, M.~Bavarian, J.~Belgum, I.~Bello, J.~Berdine, G.~Bernadett-Shapiro, C.~Berner, L.~Bogdonoff, O.~Boiko, M.~Boyd, A.-L. Brakman, G.~Brockman, T.~Brooks, M.~Brundage, K.~Button, T.~Cai, R.~Campbell, A.~Cann, B.~Carey, C.~Carlson, R.~Carmichael, B.~Chan, C.~Chang, F.~Chantzis, D.~Chen, S.~Chen, R.~Chen, J.~Chen, M.~Chen, B.~Chess, C.~Cho, C.~Chu, H.~W. Chung, D.~Cummings, J.~Currier, Y.~Dai, C.~Decareaux, T.~Degry, N.~Deutsch, D.~Deville, A.~Dhar, D.~Dohan, S.~Dowling, S.~Dunning, A.~Ecoffet, A.~Eleti, T.~Eloundou, D.~Farhi, L.~Fedus, N.~Felix, S.~P. Fishman, J.~Forte, I.~Fulford, L.~Gao, E.~Georges, C.~Gibson, V.~Goel, T.~Gogineni, G.~Goh, R.~Gontijo-Lopes, J.~Gordon, M.~Grafstein, S.~Gray, R.~Greene, J.~Gross, S.~S. Gu, Y.~Guo, C.~Hallacy, J.~Han, J.~Harris, Y.~He, M.~Heaton, J.~Heidecke, C.~Hesse, A.~Hickey,
  W.~Hickey, P.~Hoeschele, B.~Houghton, K.~Hsu, S.~Hu, X.~Hu, J.~Huizinga, S.~Jain, S.~Jain, J.~Jang, A.~Jiang, R.~Jiang, H.~Jin, D.~Jin, S.~Jomoto, B.~Jonn, H.~Jun, T.~Kaftan, Łukasz Kaiser, A.~Kamali, I.~Kanitscheider, N.~S. Keskar, T.~Khan, L.~Kilpatrick, J.~W. Kim, C.~Kim, Y.~Kim, J.~H. Kirchner, J.~Kiros, M.~Knight, D.~Kokotajlo, Łukasz Kondraciuk, A.~Kondrich, A.~Konstantinidis, K.~Kosic, G.~Krueger, V.~Kuo, M.~Lampe, I.~Lan, T.~Lee, J.~Leike, J.~Leung, D.~Levy, C.~M. Li, R.~Lim, M.~Lin, S.~Lin, M.~Litwin, T.~Lopez, R.~Lowe, P.~Lue, A.~Makanju, K.~Malfacini, S.~Manning, T.~Markov, Y.~Markovski, B.~Martin, K.~Mayer, A.~Mayne, B.~McGrew, S.~M. McKinney, C.~McLeavey, P.~McMillan, J.~McNeil, D.~Medina, A.~Mehta, J.~Menick, L.~Metz, A.~Mishchenko, P.~Mishkin, V.~Monaco, E.~Morikawa, D.~Mossing, T.~Mu, M.~Murati, O.~Murk, D.~Mély, A.~Nair, R.~Nakano, R.~Nayak, A.~Neelakantan, R.~Ngo, H.~Noh, L.~Ouyang, C.~O'Keefe, J.~Pachocki, A.~Paino, J.~Palermo, A.~Pantuliano, G.~Parascandolo, J.~Parish, E.~Parparita,
  A.~Passos, M.~Pavlov, A.~Peng, A.~Perelman, F.~de~Avila Belbute~Peres, M.~Petrov, H.~P. de~Oliveira~Pinto, Michael, Pokorny, M.~Pokrass, V.~H. Pong, T.~Powell, A.~Power, B.~Power, E.~Proehl, R.~Puri, A.~Radford, J.~Rae, A.~Ramesh, C.~Raymond, F.~Real, K.~Rimbach, C.~Ross, B.~Rotsted, H.~Roussez, N.~Ryder, M.~Saltarelli, T.~Sanders, S.~Santurkar, G.~Sastry, H.~Schmidt, D.~Schnurr, J.~Schulman, D.~Selsam, K.~Sheppard, T.~Sherbakov, J.~Shieh, S.~Shoker, P.~Shyam, S.~Sidor, E.~Sigler, M.~Simens, J.~Sitkin, K.~Slama, I.~Sohl, B.~Sokolowsky, Y.~Song, N.~Staudacher, F.~P. Such, N.~Summers, I.~Sutskever, J.~Tang, N.~Tezak, M.~B. Thompson, P.~Tillet, A.~Tootoonchian, E.~Tseng, P.~Tuggle, N.~Turley, J.~Tworek, J.~F.~C. Uribe, A.~Vallone, A.~Vijayvergiya, C.~Voss, C.~Wainwright, J.~J. Wang, A.~Wang, B.~Wang, J.~Ward, J.~Wei, C.~Weinmann, A.~Welihinda, P.~Welinder, J.~Weng, L.~Weng, M.~Wiethoff, D.~Willner, C.~Winter, S.~Wolrich, H.~Wong, L.~Workman, S.~Wu, J.~Wu, M.~Wu, K.~Xiao, T.~Xu, S.~Yoo, K.~Yu, Q.~Yuan,
  W.~Zaremba, R.~Zellers, C.~Zhang, M.~Zhang, S.~Zhao, T.~Zheng, J.~Zhuang, W.~Zhuk, and B.~Zoph.
\newblock Gpt-4 technical report, 2024.

\bibitem{oren2024transformersmultistaternns}
M.~Oren, M.~Hassid, N.~Yarden, Y.~Adi, and R.~Schwartz.
\newblock Transformers are multi-state rnns, 2024.

\bibitem{pagnoni2024bytelatenttransformerpatches}
A.~Pagnoni, R.~Pasunuru, P.~Rodriguez, J.~Nguyen, B.~Muller, M.~Li, C.~Zhou, L.~Yu, J.~Weston, L.~Zettlemoyer, G.~Ghosh, M.~Lewis, A.~Holtzman, and S.~Iyer.
\newblock Byte latent transformer: Patches scale better than tokens, 2024.

\bibitem{qwen2025qwen25technicalreport}
Qwen, :, A.~Yang, B.~Yang, B.~Zhang, B.~Hui, B.~Zheng, B.~Yu, C.~Li, D.~Liu, F.~Huang, H.~Wei, H.~Lin, J.~Yang, J.~Tu, J.~Zhang, J.~Yang, J.~Yang, J.~Zhou, J.~Lin, K.~Dang, K.~Lu, K.~Bao, K.~Yang, L.~Yu, M.~Li, M.~Xue, P.~Zhang, Q.~Zhu, R.~Men, R.~Lin, T.~Li, T.~Tang, T.~Xia, X.~Ren, X.~Ren, Y.~Fan, Y.~Su, Y.~Zhang, Y.~Wan, Y.~Liu, Z.~Cui, Z.~Zhang, and Z.~Qiu.
\newblock Qwen2.5 technical report, 2025.

\bibitem{shi2022languagemodelsmultilingualchainofthought}
F.~Shi, M.~Suzgun, M.~Freitag, X.~Wang, S.~Srivats, S.~Vosoughi, H.~W. Chung, Y.~Tay, S.~Ruder, D.~Zhou, D.~Das, and J.~Wei.
\newblock Language models are multilingual chain-of-thought reasoners, 2022.

\bibitem{shi2024discoveringgemsearlylayers}
Z.~Shi, Y.~Ming, X.-P. Nguyen, Y.~Liang, and S.~Joty.
\newblock Discovering the gems in early layers: Accelerating long-context llms with 1000x input token reduction, 2024.

\bibitem{sun2025shadowkvkvcacheshadows}
H.~Sun, L.-W. Chang, W.~Bao, S.~Zheng, N.~Zheng, X.~Liu, H.~Dong, Y.~Chi, and B.~Chen.
\newblock Shadowkv: Kv cache in shadows for high-throughput long-context llm inference, 2025.

\bibitem{tang2024questqueryawaresparsityefficient}
J.~Tang, Y.~Zhao, K.~Zhu, G.~Xiao, B.~Kasikci, and S.~Han.
\newblock Quest: Query-aware sparsity for efficient long-context llm inference, 2024.

\bibitem{lcmteam2024largeconceptmodelslanguage}
L.~team, L.~Barrault, P.-A. Duquenne, M.~Elbayad, A.~Kozhevnikov, B.~Alastruey, P.~Andrews, M.~Coria, G.~Couairon, M.~R. Costa-jussà, D.~Dale, H.~Elsahar, K.~Heffernan, J.~M. Janeiro, T.~Tran, C.~Ropers, E.~Sánchez, R.~S. Roman, A.~Mourachko, S.~Saleem, and H.~Schwenk.
\newblock Large concept models: Language modeling in a sentence representation space, 2024.

\bibitem{xia2024unlockingefficiencylargelanguage}
H.~Xia, Z.~Yang, Q.~Dong, P.~Wang, Y.~Li, T.~Ge, T.~Liu, W.~Li, and Z.~Sui.
\newblock Unlocking efficiency in large language model inference: A comprehensive survey of speculative decoding, 2024.

\bibitem{xiao2024efficientstreaminglanguagemodels}
G.~Xiao, Y.~Tian, B.~Chen, S.~Han, and M.~Lewis.
\newblock Efficient streaming language models with attention sinks, 2024.

\bibitem{xiong2024dyspecfasterspeculativedecoding}
Y.~Xiong, R.~Zhang, Y.~Li, T.~Wu, and L.~Zou.
\newblock Dyspec: Faster speculative decoding with dynamic token tree structure, 2024.

\bibitem{yao2025deftdecodingflashtreeattention}
J.~Yao, K.~Chen, K.~Zhang, J.~You, B.~Yuan, Z.~Wang, and T.~Lin.
\newblock Deft: Decoding with flash tree-attention for efficient tree-structured llm inference, 2025.

\bibitem{yu2023megabytepredictingmillionbytesequences}
L.~Yu, D.~Simig, C.~Flaherty, A.~Aghajanyan, L.~Zettlemoyer, and M.~Lewis.
\newblock Megabyte: Predicting million-byte sequences with multiscale transformers, 2023.

\bibitem{zhang2023h2oheavyhitteroracleefficient}
Z.~Zhang, Y.~Sheng, T.~Zhou, T.~Chen, L.~Zheng, R.~Cai, Z.~Song, Y.~Tian, C.~Ré, C.~Barrett, Z.~Wang, and B.~Chen.
\newblock H$_2$o: Heavy-hitter oracle for efficient generative inference of large language models, 2023.

\bibitem{zhao2024lookaheadinferenceaccelerationframework}
Y.~Zhao, Z.~Xie, C.~Liang, C.~Zhuang, and J.~Gu.
\newblock Lookahead: An inference acceleration framework for large language model with lossless generation accuracy, 2024.

\bibitem{zhou2023instructionfollowingevaluationlargelanguage}
J.~Zhou, T.~Lu, S.~Mishra, S.~Brahma, S.~Basu, Y.~Luan, D.~Zhou, and L.~Hou.
\newblock Instruction-following evaluation for large language models, 2023.

\end{thebibliography}


\newpage
\appendix

\section{Training Data Mix}

We provide a detailed training data mix consisting of all publicly accessible data in Table~\ref{tab:data}. 

\begin{table}[t]
\centering
\caption{\textbf{\ours{} Training Data Mix}: We use all public data to create the data mix for training where we select datasets spanning several domains. Mixing ratios denote percentage after sampling in terms of numbers of samples. }
\label{tab:data}
\vspace{0.2cm}
\begin{tabular}{llc}
\toprule
\textbf{Data Type} & \textbf{Dataset Name} & \textbf{Mixing Ratio} \\
\midrule
\midrule
\multirow{5}{*}{Code} & TinyCode & 3.19\% \\
 & PythonAlpaca & 3.55\% \\
 & OpenCoder & 5.97\% \\
 & OpenCoder2 & 5.85\% \\
 & MagicCoder & 5.50\% \\
\midrule
\multirow{4}{*}{General / Knowledge} & OpenOrca & 6.48\% \\
 & OpenPlatyus & 4.11\% \\
 & IFEvalLike & 4.65\% \\
 & ScienceQA & 2.47\% \\
\midrule
\multirow{7}{*}{Math} & GSM8K & 1.23\% \\
 & MetaMathQA & 6.51\% \\
 & OpenR1Math & 1.55\% \\
 & MathGPT & 3.30\% \\
 & MathInstruct & 4.32\% \\
 & OpenMathInstruct2 & 23.0\% \\
 & MathPlus & 7.37\% \\
\midrule
\multirow{3}{*}{Long Context} & NarrativeQA & 1.08\% \\
 & HotPotQA & 1.49\% \\
 & MRQA & 8.52\% \\
\bottomrule
\end{tabular}
\end{table}

\end{document}